\RequirePackage{fix-cm}
\documentclass[twocolumn]{svjour3}
\smartqed  
\usepackage{graphicx}
\usepackage{booktabs}
\usepackage{colortbl}
\usepackage{makecell}
\usepackage{tikz,pgfplots,pgfplotstable}
\usepackage{color}
\usepackage{amsfonts}
\usepackage{amssymb}
\usepackage{multirow}
\usepgfplotslibrary[groupplots]
\definecolor{mediumelectricblue}{rgb}{0.12,0.314,0.588}
\definecolor{mossgreen2}{RGB}{138,154,91}
\definecolor{internationalorange}{RGB}{100,31,0}
\definecolor{lightsalmon}{RGB}{255,160,122}

\definecolor{selfblue}{RGB}{89,138,234}
\definecolor{selfgreen}{RGB}{133,235,133}

\usepackage[pagebackref=true,breaklinks=true,colorlinks,bookmarks=false]{hyperref}

\begin{document}


\title{Unified Frequency-Assisted Transformer Framework for Detecting and Grounding Multi-Modal Manipulation
}

\author{Huan Liu \and Zichang Tan \and Qiang Chen \and Yunchao Wei \and Yao Zhao \and Jingdong Wang}

\institute{
Huan Liu, Yunchao Wei, Yao Zhao \at
Institute of Information Science, Beijing Jiaotong University. \\
Beijing Key Laboratory of Advanced Information Science and Network Technology. \\
\email{\{liu.huan,yunchao.wei,yzhao\}@bjtu.edu.cn}, 
\and
Zichang Tan, Qiang Chen, Jingdong Wang  \at
VIS, Baidu. \\
\email{\{tanzichang,chenqiang13,wangjingdong\}@baidu.com} \\
}

\date{Received: date / Accepted: date}

\maketitle

\begin{abstract}
Detecting and grounding multi-modal media manipulation (DGM$^4$) has become increasingly crucial due to the widespread dissemination of face forgery and text misinformation.
In this paper, we present the \textbf{U}nified \textbf{F}requency-\textbf{A}ssisted trans\textbf{Former} framework, named \textbf{UFAFormer}, to address the DGM$^4$ problem.
Unlike previous state-of-the-art methods that solely focus on the image (RGB) domain to describe visual forgery features, we additionally introduce the frequency domain as a complementary viewpoint. 
By leveraging the discrete wavelet transform, we decompose images into several frequency sub-bands, capturing rich face forgery artifacts. 
Then, our proposed frequency encoder, incorporating intra-band and inter-band self-attentions, explicitly aggregates forgery features within and across diverse sub-bands.
Moreover, to address the semantic conflicts between image and frequency domains, the forgery-aware mutual module is developed to further enable the effective interaction of disparate image and frequency features, resulting in aligned and comprehensive visual forgery representations.
Finally, based on visual and textual forgery features, we propose a unified decoder that comprises two symmetric cross-modal interaction modules responsible for gathering modality-specific forgery information, along with a fusing interaction module for aggregation of both modalities. 
The proposed unified decoder formulates our UFAFormer as a unified framework, ultimately simplifying the overall architecture and facilitating the optimization process. 
Experimental results on the DGM$^4$ dataset, containing several perturbations, demonstrate the superior performance of our framework compared to previous methods, setting a new benchmark in the field.

\end{abstract}

\section{Introduction}
In recent years, the Internet has witnessed 
the widely spread of fake media \cite{zheng2020survey,juefei2022countering}, 
such as face forgery images, deepfake videos, and text fake news. 
Together with the advances in deep learning,
it becomes easier to 
create hyper-realistic content,
making security and privacy a serious issue,
{\em e.g.,} face forgery for identity fraud \cite{liu2021casia,zhang2019dataset,liu2022contrastive,liu2021face} and 
text fake news for misinformation \cite{ying2023bootstrapping,zhou2023multi}.
In response to such growing threats, 
researchers have shown great attention and proposed diverse detection methods, including face forgery detection \cite{miao2023f,guandelving,miao2022hierarchical,tan2022transformer} 
and text forgery detection \cite{zhu2022generalizing,zellers2019defending} 
that focus on uni-modal ({\em i.e.,} image or text) forgery. 
Another line in previous frameworks is 
multi-modal forgery detection \cite{luo2021newsclippings,khattar2019mvae}, 
which leverages both image and text modalities and
achieves better results 
in forgery detection. 
These frameworks solely predict binary classes ({\em i.e.,} real or fake)
of given suspect inputs, 
which simply regards multi-modal forgery detection 
as a binary classification task.
More recently, the problem of {\em detecting and grounding multi-modal media manipulation} (DGM$^4$) is introduced that requires jointly \textbf{detecting and grounding} face and text manipulation \cite{shao2023detecting}.
Unlike previous binary multi-modal detection,
DGM$^4$ expands four individual sub-tasks, including 
i) binary classification for a given image-text pair,
ii) fine-grained manipulation type classification for each modality, identifying specific manipulation types such as face swap, face attribute, text swap, and text attribute,
iii) manipulated faces grounding in image, {\em i.e.,} predicting fake face bounding boxes, and 
iv) manipulated words grounding in text, {\em i.e.,} predicting fake text tokens.
As a more interpretable and challenging problem, how to address DGM$^4$ has seen significant interest.

\begin{figure}[t]
\centering
\includegraphics[width=\linewidth]{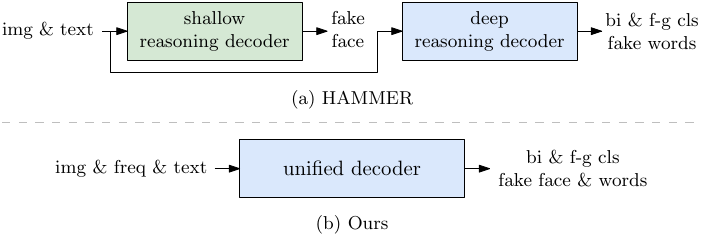}
\caption{\textbf{Comparison of transformer decoder.} Here we mainly list the overview of the decoder part between HAMMER \cite{shao2023detecting} and our UFAFormer. The difference lies in the input and decoder architecture. We build a unified decoder to model comprehensive correlation from image, text, and frequency domains, resulting in a unified framework. `img' and `freq' represent image and frequency. `bi \& f-g cls' denotes binary and fine-grained manipulation type classification.}
\label{fig1}
\end{figure}

As depicted in Figure~\ref{fig1}, the state-of-the-art approach, HAMMER \cite{shao2023detecting}, develops a multi-branch transformer structure to address DGM$^4$.
Specifically, HAMMER extracts visual and textual forgery features from image (RGB) and text domains, and adopts two dedicated decoders that separate the manipulated face grounding sub-task from other tasks.
The structure of HAMMER mainly follows recent vision-and-language representation learning \cite{li2021align,kim2021vilt,radford2021learning}, focusing solely on the interaction between image and text domains. This leads to overlooking a crucial aspect, {\em i.e.,} analyzing the frequency domain, which has been demonstrated to encompass rich face forgery artifacts in prior forgery detection studies \cite{qian2020thinking,jeong2022bihpf,miao2023f}. Intuitively, incorporating this frequency domain analysis could potentially enhance the forgery detection capabilities.
Moreover, HAMMER utilizes a multi-branch transformer structure with dedicated decoders, which contributes to the overall complexity of the architecture and may limit the framework's ability to explore the comprehensive relationships among various sub-tasks.

To address the problems, we present a novel framework called \textbf{U}nified \textbf{F}requency-\textbf{A}ssisted trans\textbf{Former}, denoted as \textbf{UFAFormer}. Our UFAFormer incorporates the frequency domain and introduces a unified transformer structure with a unified decoder, thus enabling effective detection and grounding of multi-modal media manipulation. 

\begin{figure*}[t]
\centering
\includegraphics[width=\linewidth]{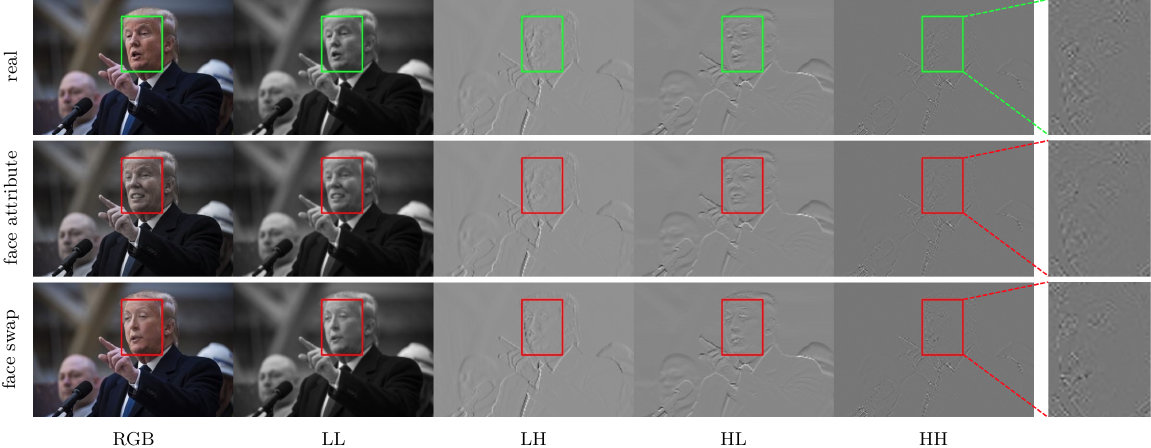}
\caption{\textbf{DWT sub-bands comparison of real and manipulated images.} We visualize two types of face manipulation methods, {\em i.e.,} face attribute \cite{gao2021information} and face swap \cite{wang2022high}. DWT transforms images into four sub-bands, including LL, LH, HL and HH, through Haar wavelet transform. Each sub-band preserves the original spatial information but has large differences in texture between real and fake. \textcolor{red}{Red} box indicates the fake face while \textcolor{green}{green} is for real. Best view in zoom in.}
\label{fig2}
\end{figure*}

Our approach is motivated by the intuition that the frequency domain can also help in tackling DGM$^4$. Hence, we consider it as a complementary viewpoint, working in conjunction with the image domain to comprehensively describe visual forgery features.
To realize this objective, we leverage the discrete wavelet transform (DWT) \cite{mallat1989theory} to decompose each input image into distinct sub-bands, including LL, LH, HL, and HH, where ``L'' and ``H'' represent low and high pass filters, respectively. As illustrated in Figure~\ref{fig2}, each DWT sub-band retains the spatial structure of the original input images and exhibits significant differences between authentic and manipulated instances. Consequently, these sub-bands serve as valuable forgery traces for detecting and grounding manipulated faces.
However, an important observation arises from the fact that forgery artifacts exhibit significant differences across different frequency sub-bands. In other words, the interaction among inter-band frequency features at different positions may not be directly helpful, as empirically confirmed in Section~\ref{sec:as}.
Based on this observation, we propose a novel frequency encoder equipped with carefully designed intra-band and inter-band self-attentions, which are tailored to explicitly aggregate both position and content forgery information from diverse sub-bands.
By considering the peculiarities of each sub-band, our approach effectively captures the essential forgery cues in the frequency domain, while avoiding potential interference from the unnecessary interaction.

After extracting frequency features, how to integrate frequency and image features together is the other important issue that needs to be studied. 
However, it is inadequate to directly incorporate them together due to the semantic conflict that arises between the image and frequency features, as expounded in Section~\ref{sec:FAMM}. 
To release this limitation, we propose an innovative component called the forgery-aware mutual module (FAMM), designed to effectively integrate these disparate features as comprehensive visual forgery features.
The primary objective of FAMM is to construct a forgery-aware feature pyramid by means of forgery-aware selection, followed by the implementation of a mutual cross-attention mechanism. 
This process involves image forgery enhancement by feature pyramid, coupled with the selection of representative forgery features. 
The cross-attention mechanism serves to bridge the domain gap between the selected image and frequency features, thereby facilitating their alignment and achieving an aligned and comprehensive representation of visual forgery features.
In essence, FAMM provides a viable solution to the semantic conflict between image and frequency features, allowing for their harmonious integration within the framework and thus leading to improved performance. 

Finally, based on visual and textual forgery features, a unified decoder is introduced to address the four sub-tasks in DGM$^4$, simplifying the architecture and facilitating the optimization process. 
In contrast to previous multi-branch methods using dedicated decoders \cite{shao2023detecting}, our unified decoder, incorporating two symmetric cross-modal interaction modules and a fusing interaction module, integrates and processes all sub-tasks simultaneously in a more streamlined and interconnected manner.
Specifically, the cross-modal interaction modules are employed to gather forgery information from diverse modalities, each with an emphasis on modality-specific manipulation, thus improving the accuracy of manipulation detecting and grounding processes.
While the fusing interaction module is responsible for aggregating forgery information across both modalities, enabling more reliable classification of the image-text pair.
In summary, the resulting decoder effectively consolidates visual and textual forgery features and represents a more integrated design, transforming our UFAFormer into a unified framework.

Experimental results show that our approach
outperforms other recent methods
on the challenging DGM$^4$ dataset.
Notably, UFAFormer achieves $93.81\%$ AUC on binary classification, $87.85\%$ AP on fine-grained manipulation type classification, $78.33\%$ IoU on manipulated face grounding, and $72.02\%$ F1 on fake text token grounding, 
setting a new state-of-the-art benchmark.
The main contributions of this paper are summarized as follows:

\begin{itemize}
    \item We present a unified and comprehensive framework for DGM$^4$, named UFAFormer. For the first time, our UFAFormer incorporates the frequency domain into the framework and simplifies its architecture with a unified decoder.
    \item A novel frequency encoder, with carefully designed intra-band and inter-band self-attentions, is proposed to explicitly aggregate both position and content forgery information from diverse sub-bands.
    \item To effectively integrate image and frequency features, we develop a forgery-aware mutual module to address the semantic conflicts, leading to aligned and comprehensive visual forgery features.
    \item We introduce a unified decoder, which simultaneously integrates and processes the detection and grounding processes for manipulated faces, text, as well as image-text pair, simplifying the architecture and facilitating the optimization process.
\end{itemize}

The rest of this paper is organized as follows. Section \ref{sec:rw} introduces relevant forgery detection and grounding methods. Section \ref{sec:method} illustrates our proposed framework for DGM$^4$. Section \ref{sec:exp} shows the experiment settings, main results, and ablation studies of UFAFormer. Section \ref{sec:conclude} concludes this paper.

\section{Related Work}\label{sec:rw}
The domain of media manipulation detection has garnered substantial attention due to the increasing prevalence of face forgery and text misinformation.
In this section, we review previous face forgery detection frameworks including spatial and frequency methods. Then, we introduce some multi-modal detection works in fake news and out-of-context misinformation. Finally, we compare recent studies related to the DGM$^4$ problem and highlight the advancements of our proposed UFAFormer.

\subsection{Face Forgery Detection}
Existing face forgery detection frameworks usually explore different forgery clues, which are roughly divided into spatial and frequency domain methods.

\noindent\textbf{Spatial-domain methods.} Most face forgery detection methods \cite{afchar2018mesonet,zhuang2022uia,guo2023controllable,wang2023altfreezing,liu2023fedforgery} extract visual forgery clues in the spatial domain. 
Early works \cite{rahmouni2017distinguishing,rossler2019faceforensics++} adopt CNN-based networks as a binary classifier to detect forgery images. 
For example, Wang {\em et al.} \cite{wang2020cnn} suggest that GAN-generated images can be surprisingly detected by a vanilla ResNet-50 \cite{he2016deep} model. Haliassos {\em et al.} \cite{haliassos2021lips} track the irregularities in mouth movements for detecting face forgery images. Besides, some works pay attention to the blending boundaries between real backgrounds and fake faces. Li {\em et al.} \cite{li2020face} and Shiohara {\em et al.} \cite{shiohara2022detecting} generate training samples by blending pseudo source and target images from real images to learn generic and robust forgery features.

\noindent\textbf{Frequency-domain methods.} There are also methods \cite{frank2020leveraging,li2021frequency,woo2022add} that aim to detect frequency artifacts, {\em e.g.,} checkerboard-like spectrum left by up-sampling, in the frequency domain.
For instance, Masi {\em et al.} \cite{masi2020two} incorporate RGB and frequency domains with a two-branch framework built by densely connected layers \cite{huang2017densely}. Qian~{\em et al.} \cite{qian2020thinking} propose F$^3$-Net also with two branches to mine forgery patterns in the image and local frequency statistics, respectively. Jeong {\em et al.} \cite{jeong2022bihpf} extract the artifact compression map and utilize it to explore frequency artifacts within Fourier transform. Besides, there are also some methods using discrete wavelet transform for face forgery detection. Jia {\em et al.} \cite{jia2021inconsistency} apply discrete wavelet transform with multiple levels to decompose RGB images and adopt HRNet \cite{wang2020deep} to extract frequency-aware features. Miao {\em et al.} \cite{miao2023f} build a frequency-based attention module by performing a discrete wavelet transform on the middle feature map.

These methods solely detect binary classes (real or fake) of given input face images, not involving multiple modalities or manipulation grounding. 
In this paper, inspired by the success of the frequency domain in face forgery detection, we undertake an exploration of its potential in the context of detecting and grounding multi-modal media manipulation.

\subsection{Multi-Modal Forgery Detection}
Several studies have been proposed for handling multi-modal forgery detection, mainly comprising fake news detection and out-of-context misinformation detection.

For fake news detection, the detection methods are expected to handle multi-modal fake news generated by humans. Jin {\em et al.} \cite{jin2017multimodal} design a multi-modal framework, consisting of an RNN module for input text and a CNN module for images, to detect fake news on Weibo and Twitter. 
Wang {\em et al.} \cite{wang2018eann} leverage an event adversarial neural network to derive consistent forgery features, benefiting the detection of fake news. Khattar {\em et al.} \cite{khattar2019mvae} introduce a multi-modal variational autoencoder coupled with a classifier for fake news detection. 
Ying {\em et al.} \cite{ying2023bootstrapping} build a multi-view detection framework to obtain the cross-modal consistency between image and text modalities. 
Zhou {\em et al.} \cite{zhou2023multi} propose a multi-grained fusion network with pre-trained BERT \cite{kenton2019bert}, CLIP \cite{radford2021learning} and Swin-T \cite{liu2021swin} to extract and fuse multi-grained features for final classification.

For out-of-context misinformation detection, an unaltered image is re-purposed to support other narratives by pairing it with a swapped text caption. 
Luo {\em et al.} \cite{luo2021newsclippings} propose the NewsCLIPpings dataset where image and text are mismatched and utilize multi-modal models for the benchmark. Abdelnabi {\em et al.} \cite{abdelnabi2022open} incorporate external image-text pairing evidence on the Web into fact-checking and automate the process with a consistency-checking network. Mu {\em et al.} \cite{mu2023self} design a self-supervised distilled strategy to supervise the learning of the detection model.

These multi-modal methods combine visual and textual information, enabling cross-modal interaction to detect manipulations in both domains. However, they primarily generate binary predictions akin to the output obtained in face forgery detection, not involving manipulation grounding.

\subsection{Multi-Modal Forgery Detection and Grounding}
Contemporary frameworks for multi-modal forgery detection and grounding are constructed based on established vision-and-language representation learning methodologies \cite{li2021align,kim2021vilt,radford2021learning}. 
These frameworks adopt hierarchical transformer structures with dedicated transformer decoders to address DGM$^4$.

\begin{figure*}[t]
\centering
\includegraphics[width=0.9\textwidth]{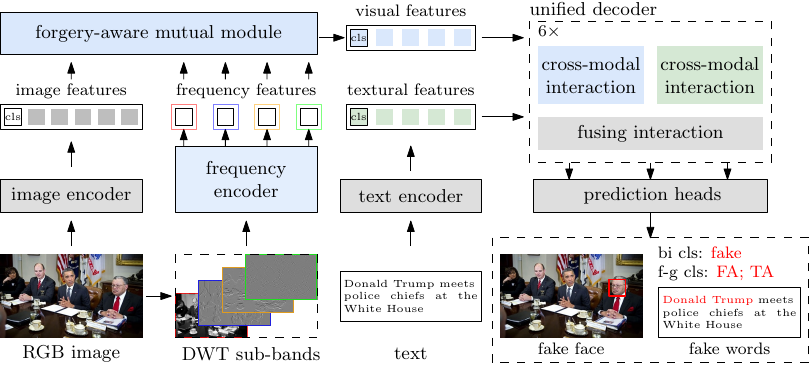}
\caption{\textbf{Our UFAFormer architecture.} There are three encoders to encode features from different domains and one unified decoder to perform interactions among them. For the image and frequency features, we further propose a forgery-aware mutual module to derive comprehensive visual forgery features with aligned semantic representation. Following the unified decoder, several prediction heads are adopted to predict the manipulation detecting and grounding results. `bi cls' and `f-g cls' denote binary and fine-grained manipulation classification. `FA' and `TA' represent face attribute and text attribute manipulation.}
\label{fig3}
\end{figure*}

Shao {\em et al.} \cite{shao2023detecting} introduce the first large-scale dataset specifically designed for the DGM$^4$ problem and present the state-of-the-art method, HAMMER, to handle it.
HAMMER leverages visual and textual forgery features extracted from the image and text domains and employs a multi-branch transformer structure with two dedicated decoders for manipulation detection and grounding.
Initially, a shallow reasoning decoder is utilized for manipulated face grounding, followed by the adoption of a deep reasoning decoder to address the remaining sub-tasks.
Despite HAMMER enabling fine-grained manipulation analysis through its multi-branch structure, it faces challenges concerning architecture complexity. 
Besides, this structure may hinder the framework's ability to fully capitalize on the comprehensive relationships between various sub-tasks, resulting in sub-optimal performance.

In contrast, our proposed UFAFormer presents two distinctive designs.
Firstly, UFAFormer not only considers the image and text domains but also integrates the frequency domain, which has been demonstrated to contain rich face forgery artifacts in forgery detection studies \cite{qian2020thinking,jeong2022bihpf,miao2023f}. By incorporating the frequency domain, our approach effectively captures additional forgery cues, resulting in improved overall performance.
Secondly, UFAFormer adopts a unified decoder that seamlessly integrates and processes the detection and grounding tasks for both image and text modalities, as well as the classification of image-text pairs. This unified decoder design streamlines the optimization process, leading to enhanced model performance.

\section{UFAFormer}\label{sec:method}
UFAFormer is a complete end-to-end unified transformer framework for detecting and grounding multi-modal media manipulation. 
Our approach incorporates the frequency domain into the framework and introduces a unified decoder for efficient processing.
In the subsequent sections, we present in detail the fundamental components of our UFAFormer.

\subsection{Overview}
The overall structure of UFAFormer is illustrated in Figure~\ref{fig3}. UFAFormer is composed of an image encoder \cite{dosovitskiyimage}, a frequency encoder, a text encoder \cite{kenton2019bert}, a forgery-aware mutual module, a unified decoder, and several task-specific prediction heads. UFAFormer is able to leverage multi-modal correlation among image, text, and frequency information to address DGM$^4$.

\noindent \textbf{Transformer encoders and decoder.}
The image encoder in our approach is implemented using $12$ layers of ViT-B/16 \cite{dosovitskiyimage}, while the text encoder is initialized with the first $6$ layers of BETR$_{\rm base}$ \cite{kenton2019bert}. For the remaining components, both the frequency encoder and unified decoder are constructed with standard transformer layers \cite{vaswani2017attention}. Each encoder processes the corresponding domain input and produces encoded uni-modal features. Subsequently, the encoded image and frequency features are directed to the forgery-aware mutual module to facilitate semantic alignment and acquire comprehensive visual forgery features. Together with the textual features, the unified decoder is utilized to perform across-modal interactions, decoupling manipulation detection and grounding processes for image, text modalities, and image-text pair.

\noindent \textbf{Prediction heads.}
As previously stated, the DGM$^4$ problem encompasses four distinct sub-tasks, including i) binary classification for image-text pairs, ii) fine-grained manipulation type classification, iii) manipulated face grounding in images, and iv) manipulated word grounding in text. Accordingly, we utilize specific prediction heads for each sub-task in UFAFormer, all implemented by feed-forward networks (FFNs).

\noindent \textbf{Loss function.}
The loss function employed in UFAFormer comprises four components: binary classification loss ($\mathcal{L}_{bi{\rm -}cls}$), fine-grained manipulation type classification loss ($\mathcal{L}_{fg{\rm -}cls}$), fake face bounding box loss for manipulated face grounding ($\mathcal{L}_{bbox}$), and fake text token loss for manipulated words grounding ($\mathcal{L}_{token}$). For the two classification losses ($\mathcal{L}_{bi{\rm -}cls}$ and $\mathcal{L}_{fg{\rm -}cls}$), we utilize binary cross entropy as the objective function. Note that $\mathcal{L}_{fg{\rm -}cls}$ represents a multi-label loss, as image-text pairs may involve both manipulated faces and words with different manipulation types.
The fake face bounding box loss ($\mathcal{L}_{bbox}$) is a combination of a normal $\ell_1$ loss and a generalized intersection over union (GIOU) loss \cite{rezatofighi2019generalized}. 
For the fake text token loss ($\mathcal{L}_{token}$), focal loss \cite{lin2017focal} with default settings ($\gamma=2,\alpha=0.25$) is employed.
We directly use the coefficients of HAMMER \cite{shao2023detecting} in loss calculation. 

\subsection{Frequency Encoder}\label{sec:FE}
\begin{figure}[t]
\centering
\includegraphics[width=\linewidth]{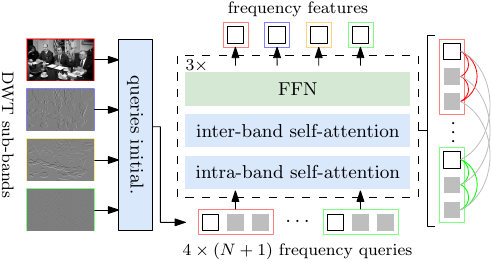}
\caption{\textbf{The structure of frequency encoder.} The encoder takes four DWT sub-bands as input and adopts queries initialization (initial.) to transform them into $4$ groups with each containing $(N+1)$ frequency queries. Then, we build intra-band and inter-band self-attentions to explicitly aggregate frequency features from different sub-bands. The hollow squares are corresponding to the sub-band embeddings.}
\label{fig4}
\end{figure}

To effectively aggregate both position and content forgery information from the frequency domain, we introduce a novel frequency encoder equipped with carefully designed intra-band and inter-band self-attentions, as depicted in Figure~\ref{fig4}. 

Although previous methods such as fast Fourier transform \cite{brigham1967fast} and discrete cosine transform \cite{rao2014discrete} can provide global forgery content information, they lack the capability to preserve position information \cite{li2020wavelet}, which is crucial in the context of DGM$^4$ for accurate localization during the manipulated face grounding sub-task.
To address this limitation, we adopt the discrete wavelet transform (DWT) \cite{mallat1989theory}, which retains the spatial structure of the original image while simultaneously capturing discriminative frequency forgery content details.
We utilize DWT to decompose a given image into four distinct sub-bands, {\em i.e.,} LL, LH, HL, and HH, as shown in Figure~\ref{fig2}. Here, ``L" and ``H" refer to low and high pass filters, capturing smooth surfaces and uneven edges, respectively. The derived DWT sub-bands contain different frequency information retaining subtle artifacts such as blending boundaries. 
In previous work \cite{miao2023f}, these sub-bands are directly concatenated (along the channel dimension) to form a mixed image, followed by the standard self-attentions \cite{dosovitskiyimage} for encoding.
However, we observe that forgery artifacts manifest differently in each frequency sub-band. This implies that the interaction among inter-band frequency features of different positions may not be directly helpful (empirically confirmed in Section~\ref{sec:as}).
To address this issue, we propose a more effective approach within our frequency encoder. Specifically, we introduce four learnable sub-band embeddings, each initialized randomly and corresponding to one of the sub-bands. These embeddings enable adaptive aggregation using two subsequent self-attentions, the intra-band self-attentions and inter-band self-attentions, while removing the unnecessary interaction.

\noindent \textbf{Queries initialization.}
Given four DTW sub-bands ($4 \times \mathbb{R}^{H\times W}$), with height $H$ and width $W$, we adopt patch embedding \cite{dosovitskiyimage} to transform and stack them into flattened 3D patches sized $\mathbb{R}^{4\times N\times P^2}$. Here, $P$ denotes the patch size, and $N=\frac{HW}{P^2}$ represents the number of patches in each sub-band.
To ensure consistency with the constant latent vector size $D$ in UFAFormer, we apply a linear projection to map these patches to $D$ dimensions, resulting in frequency embeddings of shape $\mathbb{R}^{4\times N\times D}$. Subsequently, based on these frequency embeddings, we initialize frequency queries, finally yielding the desired frequency features.

To explicitly aggregate both position and content forgery information, each frequency query in UFAFormer is structured with content and position parts. 
In detail, we prepend the $4$ randomly initialized learnable sub-band embeddings ($\mathbb{R}^{4\times D}$) to the frequency embeddings ($\mathbb{R}^{4\times N\times D}$) to yield the content part ($\mathbb{R}^{4\times (N+1)\times D}$) of frequency queries. 
This incorporation is essential, as the content information of each frequency query is closely related to the local frequency features within the corresponding patch, with the additional embeddings serving the purpose of adaptive aggregation.
For the position part, standard learnable 1D position embeddings ($\mathbb{R}^{(N+1)\times D}$) are employed to preserve the spatial position information, aligning with the image encoder \cite{dosovitskiyimage}.
Finally, to form frequency queries ($\mathbb{R}^{4\times (N+1)\times D}$), the content and position parts are merged by simply adding them together, enabling the capture of both position and content forgery information.

\noindent \textbf{Intra-band self-attentions.}
We build interactions among the frequency queries within each sub-band. It calculates $4$ self-attention maps in parallel with shapes $(N+1) \times (N+1)$, extracting forgery features in each specific sub-band (as indicated by the \textcolor{red}{red} and \textcolor{green}{green} curves in Figure~\ref{fig4}). Here, we omit the batch size and attention heads.

\noindent \textbf{Inter-band self-attentions.}
Similar to the intra-band self-attentions, we further build interaction across different sub-bands, exchanging and integrating forgery information from queries across other sub-bands with the same spatial position (illustrated by the \textcolor{gray}{gray} curves in Figure~\ref{fig4}.). As a result, UFAFormer generates $(N+1)$ inter-band self-attention maps in parallel, each with a shape of $4 \times 4$.

In contrast to standard self-attentions \cite{vaswani2017attention}, the devised intra-band and inter-band self-attentions in UFAFormer are specifically tailored to explicitly aggregate both content and position forgery information from queries within the same sub-band and across different sub-bands of the same position. 
Notably, the inter-band interaction for queries of different positions is deliberately removed, to avoid potential interference and ensure the effective utilization of forgery information from the frequency domain.
Subsequently, the resulting aggregated frequency features ($\mathbb{R}^{4\times D}$), which correspond to the four prepended sub-band embeddings, are then fed into the following forgery-aware mutual module. 

\subsection{Forgery-Aware Mutual Module}\label{sec:FAMM}
\begin{figure}[t]
\centering
\includegraphics[width=\linewidth]{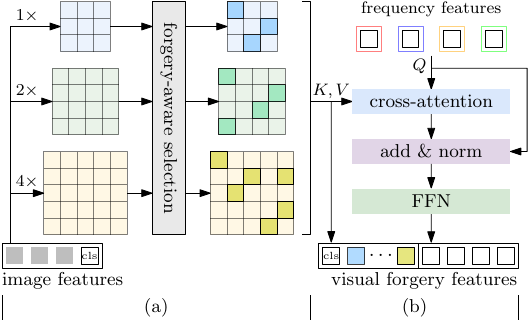}
\caption{\textbf{The structure of forgery-aware mutual module (FAMM).} FAMM consists of (a) forgery-aware feature pyramid and (b) mutual cross-attention. We first perform the forgery-aware feature pyramid to sample and gather representative multi-scale image features. Then, mutual cross-attention is adopted to mix with frequency features and derive the comprehensive visual forgery features. Here we omit the $0.5\times$ feature map without selection. Best view in color.}
\label{fig5}
\end{figure}

After extracting frequency features, the other critical challenge is how to integrate with the existing image features. 
However, direct incorporating these disparate features is insufficient due to the semantic conflict that arises between image and frequency features.
The underlying semantic conflict stems from two key aspects. Firstly, the relatively small size of manipulated faces in most images often results in encoded image features mainly representing authentic backgrounds, while the frequency features contain aggregated forgery-related information. Secondly, a natural domain gap exists between the image and frequency domains.
To address these issues, we present an innovative forgery-aware mutual module (FAMM), consisting of the forgery-aware feature pyramid (depicted in Figure~\ref{fig5} (a)) and mutual cross-attention mechanism (illustrated in Figure~\ref{fig5} (b)).
These two modules in FAMM mitigate the semantic conflict, thereby reconciling and combining the image and frequency features into aligned and comprehensive visual forgery features.

\noindent \textbf{Forgery-aware feature pyramid.}
In order to emphasize manipulated face regions and obtain forgery-related image features, a forgery-aware feature pyramid is employed in FAMM.
Inspired by \cite{li2022exploring}, we establish a differentiable feature pyramid by adopting $4$ convolution operations with strides $\{ 2, 1, \frac{1}{2}, \frac{1}{4}\}$, thereby generating multi-scale feature maps through the reshaped image features. Here, fractional stride correspond to deconvolution operation. 
By magnifying the presence of forgery-related features associated with manipulated faces through this multi-scale feature representation, the forgery features can be prominently enhanced within the forgery-aware feature pyramid.
However, during the feature pyramid process, the features of authentic backgrounds are also enhanced, as it inherently lacks the ability to autonomously identify forgery-related features.

To address this problem and focus on discerning image manipulation, a novel approach termed forgery-aware selection is introduced. Specifically, binary forgery classification and manipulated face grounding heads, both implemented using feed-forward networks (FFNs), are attached to the generated multi-scale feature maps. These heads collaboratively predict forgery-related scores and fake face bounding boxes for each location within the feature maps.
During training, Hungarian matching \cite{carion2020end} is adopted for supervision to match the predicted outcomes with ground truth annotations. Thus, a higher score signifies a more precise prediction regarding face detecting and grounding. Leveraging the predicted forgery-related scores, a set of sampling locations is determined for each scale of feature maps. 
Finally, the forgery-related features can be gathered through the sampling of multi-scale image forgery feature maps, ensuring the FAMM focusing on forgery-related information.

\noindent \textbf{Mutual cross-attention.}
To mitigate the domain gap between image and frequency domains, we adopt a simple yet effective mutual cross-attention to further align their representation. Inspired by the multi-head attention \cite{vaswani2017attention}, we utilize the frequency features as the guidance to engage in information exchange with the gathered forgery-related features via cross-attention interactions. Subsequently, the aligned frequency features and the gathered forgery-related features are concatenated to derive the aligned and comprehensive visual forgery features, serving as input to the following unified decoder.

\subsection{Unified Decoder}
\begin{figure}[t]
\centering
\includegraphics[width=0.9\linewidth]{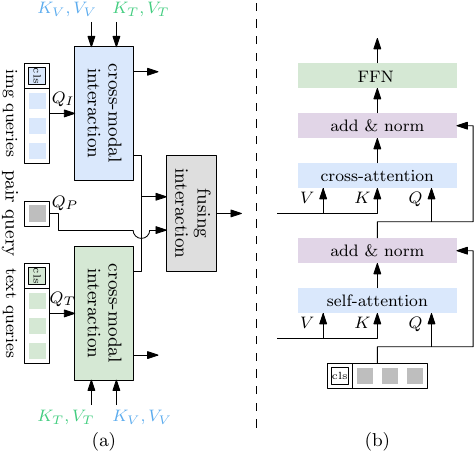}
\caption{\textbf{The structure of (a) the unified decoder and (b) the cross-modal interaction module.} Note that we conduct the manipulation detection and grounding processes in a unified manner. The subscripts $_V$ and $_T$ of $K,V$ denote the visual and textual forgery features, while $_I$, $_P$ and $_T$ of $Q$ represent the corresponding image, image-text pair and text forgery queries, respectively.}
\label{fig6}
\end{figure}

Based on the visual forgery features from FAMM and textual forgery features from the text encoder, we introduce a unified decoder to simultaneously integrate and process the four sub-tasks in DGM4, simplifying the architecture and facilitating the optimization process.
Specifically, the proposed unified decoder consists of two symmetric cross-modal interaction modules, gathering cross-modal inconsistency for modality-specific manipulation detection and grounding, and a fusing interaction module for the image-text pair binary classification, as illustrated in Figure~\ref{fig6} (a). 
In contrast to the multi-branch structure with dedicated decoders \cite{shao2023detecting}, the proposed unified decoder is able to consolidate visual and textual forgery features and capitalize on the comprehensive relationships among various sub-tasks in DGM$^4$, formulating our UFAFormer as a unified framework.

\noindent \textbf{Queries initialization.}
There are three types of queries in the proposed unified decoder, including image forgery queries, text forgery queries, and an image-text pair forgery query. The former two query types are dedicated to modality-specific manipulation detection and grounding processes, while the third is used for the binary classification of the given image-text pair.

We construct the image forgery queries ($\mathbb{R}^{(1+K)\times D}$) by concatenating image \texttt{CLS} token ($\mathbb{R}^{1\times D}$) from the image encoder and $K$ randomly initialized learnable grounding embeddings ($\mathbb{R}^{K\times D}$). 
Particularly, the \texttt{CLS} token is utilized for the fine-grained manipulation type classification of manipulated faces, as it has encapsulated global information of the input image during the image encoding \cite{dosovitskiyimage}.
To initialize text forgery queries, we directly use textual features sized $\mathbb{R}^{(1+L) \times D}$ , where $1$ and $L$ represent the text \texttt{CLS} token and encoded text features with a sequence length of $L$ from the text encoder, respectively. 
Similarly, the text \texttt{CLS} token serves text manipulation fine-grained classification, while the remaining sequence is employed for manipulated word grounding. 
Here, we do not utilize randomly initialized embedding for manipulated word grounding,  considering the varying text lengths $L$ among different samples.
For the image-text pair forgery query ($\mathbb{R}^{1\times D}$), we adopt a randomly initialized learnable embedding.

\noindent \textbf{Cross-modal interaction module.}
There are two symmetric cross-modal interaction modules, each with an emphasis on modality-specific manipulation, in our unified decoder.
As depicted in Figure~\ref{fig6} (b), this module builds cross-modal interactions with the visual and textual forgery features, as well as the corresponding forgery queries.
As an illustrative example, we describe the cross-modal interaction process for image forgery queries, as follows.

Given the image forgery queries, we first perform self-attention with visual forgery features from FAMM, exploring and aggregating the inconsistency in the visual features. Then, cross-attention is used to further exchange forgery information with the textual features derived from the text encoder, focusing on capturing inconsistency across the visual and textual modalities. At last, we adopt FFN to obtain the final cross-modal forgery features, sent to corresponding detecting and grounding heads. The steps of processing text forgery queries are similar and omitted for simplicity.

\noindent \textbf{Fusing interaction module.}
Connected with the above two cross-modal interaction modules, the proposed fusing interaction module is used for the binary classification of the image-text pair.
Specifically, we adopt a standard multi-head attention module \cite{vaswani2017attention} to perform the fusing interaction between the image-text pair forgery query and the cross-modal interaction module outputs, which is followed by an FFN for classification.
In this way, this module aggregates forgery information across both visual and textual aspects, enabling more reliable classification of the image-text pair.

\section{Experiments} \label{sec:exp}
\subsection{Settings}
\noindent \textbf{Dataset.} We conduct our experiments on the DGM$^4$ dataset \cite{shao2023detecting}, containing $230$K image-text paired samples with over $77$K pristine pairs and $152$K manipulated pairs. There are four types of manipulation in this dataset, including face swap (FS), face attribute (FA), text swap (TS), and text attribute (TA). The DGM$^4$ dataset is challenging as the manipulated face and manipulated text are randomly combined together. Moreover, several perturbations, {\em e.g.,} JPEG compression and Gaussian noise, are employed on half of the whole dataset, making it closer to the real world scenario. We train UFAFormer on the \texttt{train} set and evaluate it on \texttt{val} set and \texttt{test} set.

\noindent \textbf{Evaluation metric.} We report our results following the original evaluation protocols and metrics \cite{shao2023detecting}. Here, we detail the metrics for each DGM$^4$ sub-task. For \textit{binary classification} of image-text pair, we adopt accuracy (ACC), area under the receiver operating characteristic curve (AUC), and equal error rate (EER). For \textit{fine-grained manipulation type classification}, the mean average precision (mAP), average per-class F1 (CF1), and average overall F1 (OF1) are used. For \textit{manipulated faces grounding} in image, the mean of intersection over union (IoU$\rm _m$) and IoU with two thresholds $\{0.5, 0.75\}$ are considered, denoted as IoU$\rm _{50}$ and IoU$\rm _{75}$. For \textit{manipulated words grounding} in text, precision (PR), Recall (RE), and F1 score are employed, due to the class imbalance scenario.

\noindent \textbf{Implementation details.} For fair comparisons, our training configuration follows HAMMER \cite{shao2023detecting}. Several random image augmentations, including auto-contrast, equalize, brightness and sharpness, and random flip, are applied during training. The input images are resized into $256\times 256$, and the text sequence is padded with a max length of $50$. The AdamW \cite{kingma2014adam,loshchilov2017decoupled} is adopted with a weight decay of $0.02$. We set the base learning rate as $2\times 10^{-5}$ under a cosine schedule and train UFAFormer with $50$ epochs.

\begin{table*}[ht]
\renewcommand{\arraystretch}{1.30}
\caption{\textbf{Comparisons with state-of-the-art methods on DGM$^4$ dataset.} We report the results on \texttt{test} set by default, except for $\dagger$ representing \texttt{val} set. For uni-modal methods, the evaluation is conducted on the image (img) or text subset (sub.). Results of some other methods are cited from \cite{shao2023detecting}. The `PR' and `RE' denote precision and recall. $\downarrow$ denotes less is better. The \textbf{best results} are highlighted in \textbf{bold}.}\label{tab1}
\resizebox{\textwidth}{!}{
\begin{tabular}{clccccccccccccc}
\toprule
& & & \multicolumn{3}{@{}c@{}}{Binary Class} & \multicolumn{3}{@{}c@{}}{Fine-Grained Class} & \multicolumn{3}{@{}c@{}}{Image Grounding} & \multicolumn{3}{@{}c@{}}{Text Grounding} \\\cmidrule{4-15}
& Methods & Type & AUC & EER$\downarrow$ & ACC & mAP & CF1 & OF1 & IoU$_{\rm m}$ & IoU$_{50}$ & IoU$_{75}$ & PR & RE & F1 \\\midrule
\multirow{3}{*}{\rotatebox{90}{img sub.}}&TS \cite{luo2021generalizing} & image & $91.80$ & $17.11$ & $82.89$ & - & - & - & $72.85$ & $79.12$ & $74.06$ & - & - & - \\
&MAT \cite{zhao2021multi} & image & $91.31$ & $17.65$ & $82.36$ & - & - & - & $72.88$ & $78.98$ & $74.70$ & - & - & - \\
&UFAFormer & multi-modal  & $\mathbf{94.88}$ & $\mathbf{12.35}$ & $\mathbf{87.16}$ & - & - & - & $\mathbf{77.28}$ & $\mathbf{85.46}$ & $\mathbf{78.29}$ & - & - & - \\
\midrule
\multirow{3}{*}{\rotatebox{90}{text sub.}}&BETR \cite{kenton2019bert} & text & $80.82$ & $28.02$ & $68.98$ & - & - & - & - & - & - & $41.39$ & $63.85$ & $50.23$ \\
&LUKE \cite{yamada2020luke} & text & $81.39$ & $27.88$ & $76.18$ & - & - & - & - & - & - & $50.52$ & $37.93$ & $43.33$ \\
&UFAFormer & multi-modal  & $\mathbf{94.11}$ & $\mathbf{12.61}$ & $\mathbf{84.71}$ & - & - & - & - & - & - & $\mathbf{81.13}$ & $\mathbf{70.73}$ & $\mathbf{75.58}$ \\
\midrule
\multirow{7}{*}{\rotatebox{90}{entire dataset}}&CLIP \cite{radford2021learning} & multi-modal & $83.22$ & $24.61$ & $76.40$ & $66.00$ & $59.52$ & $62.31$ & $49.51$ & $50.03$ & $38.79$ & $58.12$ & $22.11$ & $32.03$ \\
&ViLT \cite{kim2021vilt} & multi-modal & $85.16$ & $22.88$ & $78.38$ & $72.37$ & $66.14$ & $66.00$ & $59.32$ & $65.18$ & $48.10$ & $66.48$ & $49.88$ & $57.00$ \\
&HAMMER$^\dagger$ \cite{shao2023detecting} & multi-modal & $92.71$ & $14.26$ & $85.72$ & $85.21$ & $78.30$ & $79.59$ & $75.72$ & $83.30$ & $75.23$ & $73.45$ & $64.79$ & $68.85$ \\
&HAMMER \cite{shao2023detecting} & multi-modal & $93.19$ & $14.10$ & $86.39$ & $86.22$ & $79.37$ & $80.37$ & $76.45$ & $83.75$ & $76.06$ & $\mathbf{75.01}$ & $68.02$ & $71.35$ \\
\cmidrule{2-15}
&\cellcolor{lightgray!20}{UFAFormer$^\dagger$} & \cellcolor{lightgray!20}{multi-modal} & \cellcolor{lightgray!20}{$93.41$} & \cellcolor{lightgray!20}{$14.20$} & \cellcolor{lightgray!20}{$86.50$} & \cellcolor{lightgray!20}{$86.98$} & \cellcolor{lightgray!20}{$79.55$} & \cellcolor{lightgray!20}{$81.05$} & \cellcolor{lightgray!20}{$77.82$} & \cellcolor{lightgray!20}{$85.15$}& \cellcolor{lightgray!20}{$78.22$} & \cellcolor{lightgray!20}{$71.72$} & \cellcolor{lightgray!20}{$68.99$} & \cellcolor{lightgray!20}{$70.33$} \\
&\cellcolor{lightgray!20}{UFAFormer} & \cellcolor{lightgray!20}{multi-modal} & \cellcolor{lightgray!20}{$\mathbf{93.81}$} & \cellcolor{lightgray!20}{$\mathbf{13.60}$} & \cellcolor{lightgray!20}{$\mathbf{86.80}$} & \cellcolor{lightgray!20}{$\mathbf{87.85}$} & \cellcolor{lightgray!20}{$\mathbf{80.31}$} & \cellcolor{lightgray!20}{$\mathbf{81.48}$} & \cellcolor{lightgray!20}{$\mathbf{78.33}$} & \cellcolor{lightgray!20}{$\mathbf{85.39}$} & \cellcolor{lightgray!20}{$\mathbf{79.20}$} & \cellcolor{lightgray!20}{$73.35$} & \cellcolor{lightgray!20}{$\mathbf{70.73}$} & \cellcolor{lightgray!20}{$\mathbf{72.02}$} \\
\bottomrule
\end{tabular}}
\end{table*}

\subsection{Main Results}
In this paper, we aim to address the problem of detecting and grounding multi-modal media manipulation. Thus, we mainly compare UFAFormer with previous multi-modal frameworks, including CLIP \cite{radford2021learning}, ViLT \cite{kim2021vilt}, and HAMMER \cite{shao2023detecting}. Besides, to show the effectiveness of multi-modal learning, we also consider the comparison with some representative uni-modal frameworks, such as image-based methods \cite{luo2021generalizing,zhao2021multi} and text-based methods \cite{kenton2019bert,yamada2020luke}.

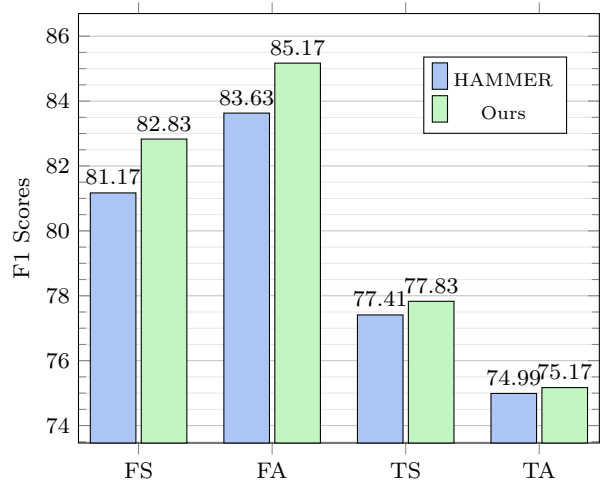
\begin{figure}
\centering
\begin{tikzpicture}[font=\footnotesize]
\begin{axis}[
    ymajorgrids,
    yminorgrids,
    grid style={line width=.1pt, draw=gray!20},
    major grid style={line width=.2pt,draw=gray!50},
    minor y tick num=3,
    ybar,
    enlargelimits=0.15,
    legend style={
    at={(0.8,0.9)},
      anchor=north,legend columns=1,font=\scriptsize},
    legend image code/.code={
        \draw [#1] (0cm,-0.1cm) rectangle (0.2cm,0.25cm); },
    ylabel={F1 Scores},
    y label style={at={(0.07,0.5)}},
    symbolic x coords={FS,FA,TS,TA},
    xtick=data,
    nodes near coords,
    nodes near coords align={vertical},
    ]
\addplot[bar width=0.6cm,fill=selfblue!50] coordinates {(FS,81.17) (FA,83.63) (TS,77.41) (TA,74.99)};
\addlegendentry{HAMMER}
\addplot[bar width=0.6cm,fill=selfgreen!50] coordinates {(FS,82.83) (FA,85.17) (TS,77.83) (TA,75.17)};
\addlegendentry{Ours}
\end{axis}
\end{tikzpicture}
\caption{\textbf{F1 scores of four different manipulation types in fine-grained manipulation type classification.} We report the results on DGM$^4$ \texttt{test} set. Our UFAFormer works better on all manipulation types, including face swap (FS), face attribute (FA), text swap (TS), and text attribute (TA).}
\label{fig7}
\end{figure}

\noindent\textbf{Comparisons with multi-modal frameworks.}
Table~\ref{tab1} presents the detailed results of four sub-tasks in the DGM$^4$ dataset. Results show that our proposed UFAFormer outperforms CLIP \cite{radford2021learning}, ViLT \cite{kim2021vilt}, and HAMMER \cite{shao2023detecting} consistently. 

On DGM$^4$ \texttt{test} set, UFAFormer achieves $93.81\%$ AUC, $87.85\%$ mAP, $78.33\%$ IoU$\rm _{m}$, and $72.02\%$ F1 on binary classification, fine-grained manipulation type classification, manipulated faces grounding, and manipulated words grounding sub-tasks, respectively. UFAFormer exceeds CLIP and ViKT by a large margin of over $10\%$ on most evaluation metrics, demonstrating the effectiveness of our proposed framework for DGM$^4$. 
When compared with the recently proposed HAMMER, which adopts multi-branch structure with dedicated decoders, UFAFormer also surpasses it with non-negligible improvements, such as $+1.63\%$ mAP on fine-grained manipulation type classification and $+1.88\%$ IoU$\rm _{m}$ on manipulated faces grounding. 
Although our precision of manipulated words grounding is worse than HAMMER, UFAFormer obtains a much better recall score of $70.73\%$. This is mainly because of the class imbalance in this sub-task, where manipulated words are much fewer than real ones. Therefore, we recommend the F1 score as a more balance metric, getting $+0.67\%$ gains than HAMMER, demonstrating that our approach can obtain a better trade-off on this sub-task.
Besides, for the fine-grained manipulation type classification, Figure~\ref{fig7} illustrates the detailed F1 scores of four different manipulation types in the DGM$^4$ dataset. We can observe that UFAFormer surpasses HAMMER in all manipulation types, especially for the face manipulation, {\em e.g., } $+1.66\%$ on face swap and $+1.54\%$ on face attribute.
We also see that the type of text manipulation is harder to be identified, but we still achieve a slightly better performance than HAMMER. The above evidence indicates that the design of multi-branch structure like HAMMER may not be necessary for addressing the DGM$^4$ problem. On DGM$^4$ \texttt{val} set, our UFAFormer also achieves better performance compared with HAMMER, similar to the trends observed in DGM$^4$ \texttt{test} set.

\begin{figure*}[t]
\centering
\includegraphics[width=0.95\textwidth]{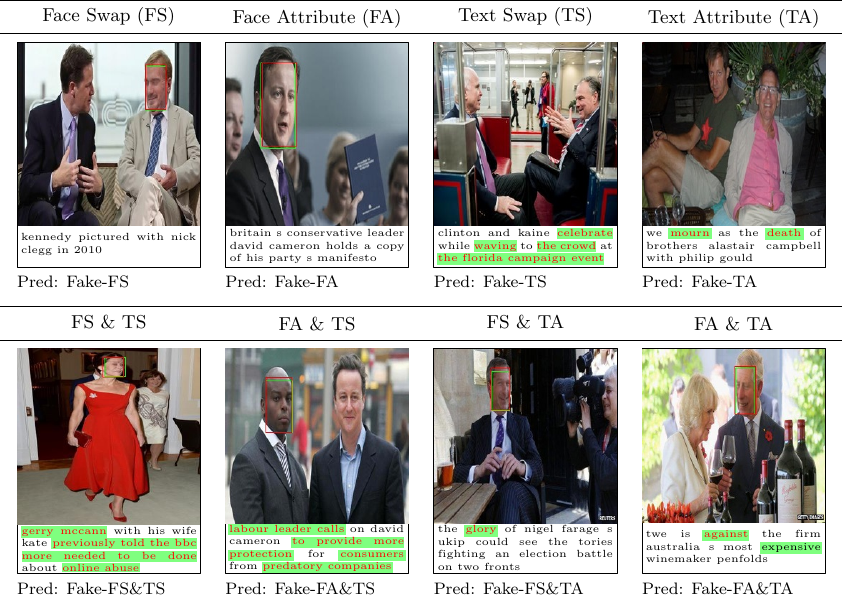}
\caption{\textbf{Visualization of detection and grounding results on different manipulation types.} UFAFormer performs well on single uni-modal manipulation and mixed multi-modal manipulation, giving precise detecting and grounding results. \textcolor{red}{Red box and text} indicate the prediction of manipulated faces and words, while \textcolor{green}{green} corresponds to ground truth annotations. Best view in color.}
\label{fig9}
\end{figure*}

\noindent\textbf{Comparisons with uni-modal frameworks.}
We compare UFAFormer with some uni-modal frameworks, including two representative face forgery detection methods \cite{luo2021generalizing,zhao2021multi} with appended grounding heads and two widely-used NLP methods \cite{kenton2019bert,yamada2020luke} for word sequence tagging. Table~\ref{tab1} shows that these uni-modal frameworks commonly achieve unsatisfactory manipulation grounding performance, especially for text manipulation grounding by an over $-20\%$ F1 gap. We think this phenomenon is caused by the differences in the learning procedure, {\em e.g.,} text-based frameworks only learn the inconsistency inside the given text sequence, which challenges manipulated word grounding. In contrast, our UFAFormer, which further leverages the inconsistency between image and text modalities, can learn more comprehensive representation and thus achieves better results than uni-modal methods.

\noindent\textbf{Visualization of detection and grounding results.} Figure~\ref{fig9} illustrates the visualization of some test samples with different manipulation types. The first row shows that our UFAFormer can effectively detect and ground forgery media, each containing a single uni-modal manipulation. Besides, under a more challenging mixed multi-modal manipulation scenario, where both manipulated faces and words exist, UFAFormer still achieves good performance and identifies most forgery parts. The results visually demonstrate the effectiveness of our UFAFormer in detecting and grounding multi-modal media manipulation.

\begin{table*}[t]
\caption{\textbf{Ablation on each proposed component.} Each component, including the frequency (freq.) encoder, forgery-aware mutual module (FAMM), and unified decoder, in UFAFormer is necessary to achieve good performance. Without (w/o) any of them leads to a performance drop.}\label{ablation:0}
\resizebox{\textwidth}{!}{
\begin{tabular}{ccccccccccccc}
\toprule
\multirow{2}{*}{\makecell{UFAFormer\\Components}} & \multicolumn{3}{@{}c@{}}{Binary Class} & \multicolumn{3}{@{}c@{}}{Fine-Grained Class} & \multicolumn{3}{@{}c@{}}{Image Grounding} & \multicolumn{3}{@{}c@{}}{Text Grounding} \\\cmidrule{2-4}\cmidrule{5-7}\cmidrule{8-10}\cmidrule{11-13}
& AUC & EER$\downarrow$ & ACC & mAP & CF1 & OF1 & IoU$_{\rm m}$ & IoU$_{50}$ & IoU$_{75}$ & PR & RE & F1 \\\midrule
w/o freq. encoder & $92.92$ & $14.75$ & $86.02$ & $86.59$ & $78.46$ & $79.96$ & $76.33$ & $83.74$ & $76.89$ & $69.89$ & $67.65$ & $68.75$ \\
w/o FAMM & $92.79$ & $14.91$ & $85.60$ & $86.20$ & $78.01$ & $79.39$ & $74.72$ & $82.86$ & $71.25$ & $\mathbf{72.14}$ & $66.24$ & $69.06$ \\
w/o unified decoder & $92.33$ & $14.64$ & $85.57$ & $85.13$ & $77.97$ & $79.05$ & $75.60$ & $83.40$ & $73.94$ & $71.19$ & $67.36$ & $69.22$ \\
\midrule
\cellcolor{lightgray!20}{UFAFormer} & \cellcolor{lightgray!20}{$\mathbf{93.41}$} & \cellcolor{lightgray!20}{$\mathbf{14.20}$} & \cellcolor{lightgray!20}{$\mathbf{86.50}$} & \cellcolor{lightgray!20}{$\mathbf{86.98}$} & \cellcolor{lightgray!20}{$\mathbf{79.55}$} & \cellcolor{lightgray!20}{$\mathbf{81.05}$} & \cellcolor{lightgray!20}{$\mathbf{77.82}$} & \cellcolor{lightgray!20}{$\mathbf{85.15}$} & \cellcolor{lightgray!20}{$\mathbf{78.22}$} & \cellcolor{lightgray!20}{${71.72}$} & \cellcolor{lightgray!20}{$\mathbf{68.99}$} & \cellcolor{lightgray!20}{$\mathbf{70.33}$} \\
\bottomrule
\end{tabular}}
\end{table*}

\begin{table*}[t]
\caption{\textbf{Frequency encoder implementations.} Both the intra-band and inter-band self-attentions are important in the frequency (freq.) encoder implementation. Besides, the forgery aggregation by prepending sub-band embedding also boosts performance.}\label{ablation:1}
\resizebox{\textwidth}{!}{
\begin{tabular}{cccccccc}
\toprule
\multirow{2}{*}{Freq. Encoder Implementation} & \multirow{2}{*}{Forgery Aggregation} & \multicolumn{3}{@{}c@{}}{Image Grounding} & \multicolumn{3}{@{}c@{}}{Text Grounding} \\ \cmidrule{3-8}
& & IoU$_{\rm m}$ & IoU$_{50}$ & IoU$_{75}$ & PR & RE & F1 \\
\midrule
standard self-attentions   & \checkmark & $76.78$ & $84.72$ & $76.12$ & $71.13$ & $67.82$ & $69.43$ \\
intra-band self-attentions & \checkmark & $75.48$ & $83.49$ & $73.63$ & $69.51$ & $68.49$ & $69.00$ \\
inter-band self-attentions & \checkmark & $73.62$ & $81.49$ & $71.72$ & $71.09$ & $63.84$ & $67.27$ \\
intra-band \& inter-band self-attentions & $\times$ & $75.83$ & $83.59$ & $73.98$ & $70.91$ & $68.43$ & $69.65$ \\
\cellcolor{lightgray!20}{intra-band \& inter-band self-attentions} & \cellcolor{lightgray!20}{\checkmark} & \cellcolor{lightgray!20}{$\mathbf{77.82}$} & \cellcolor{lightgray!20}{$\mathbf{85.15}$} & \cellcolor{lightgray!20}{$\mathbf{78.22}$} & \cellcolor{lightgray!20}{$\mathbf{71.72}$} & \cellcolor{lightgray!20}{$\mathbf{68.99}$} & \cellcolor{lightgray!20}{$\mathbf{70.33}$} \\
\bottomrule
\end{tabular}}
\end{table*}

\subsection{Ablation Study}\label{sec:as}
In this section, we conduct a number of ablation experiments to study the effects of key elements and hyper-parameters in UFAFormer. Unless specified, we report the results on DGM$^4$ \texttt{val} set and mark the default settings with \colorbox{lightgray!20}{gray} in the ablation tables.

\begin{table*}[t]
\renewcommand{\arraystretch}{1.2}
\caption{\textbf{Efficacy of forgery-aware mutual module.} The proposed forgery-aware feature pyramid (FAFP) and mutual cross-attention (MCA) are both essential.}\label{ablation:3}
\resizebox{\textwidth}{!}{
\begin{tabular}{cccccccccccccc}
\toprule
& & \multicolumn{3}{@{}c@{}}{Binary Class} & \multicolumn{3}{@{}c@{}}{Fine-Grained Class} & \multicolumn{3}{@{}c@{}}{Image Grounding} & \multicolumn{3}{@{}c@{}}{Text Grounding} \\\cmidrule{3-14}
FAFP & MCA & AUC & EER$\downarrow$ & ACC & mAP & CF1 & OF1 & IoU$_{\rm m}$ & IoU$_{50}$ & IoU$_{75}$ & PR & RE & F1 \\\midrule
& & $92.79$ & $14.91$ & $85.60$ & $86.20$ & $78.01$ & $79.39$ & $74.72$ & $82.86$ & $71.25$ & $\mathbf{72.14}$ & $66.24$ & $69.06$ \\
\checkmark &  & $92.91$ & $14.69$ & $85.93$ & $86.35$ & $78.61$ & $80.04$ & $76.57$ & $83.82$ & $76.87$ & $71.27$ & $68.19$ & $69.70$ \\
& \checkmark & $93.07$ & $14.55$ & $86.12$ & $86.46$ & $78.70$ & $80.13$ & $76.08$ & $83.79$ & $74.68$ & $71.45$ & $67.24$ & $69.28$ \\
\cellcolor{lightgray!20}{\checkmark} & \cellcolor{lightgray!20}{\checkmark} & \cellcolor{lightgray!20}{$\mathbf{93.41}$} & \cellcolor{lightgray!20}{$\mathbf{14.20}$} & \cellcolor{lightgray!20}{$\mathbf{86.50}$} & \cellcolor{lightgray!20}{$\mathbf{86.98}$} & \cellcolor{lightgray!20}{$\mathbf{79.55}$} & \cellcolor{lightgray!20}{$\mathbf{81.05}$} & \cellcolor{lightgray!20}{$\mathbf{77.82}$} & \cellcolor{lightgray!20}{$\mathbf{85.15}$} & \cellcolor{lightgray!20}{$\mathbf{78.22}$} & \cellcolor{lightgray!20}{$71.72$} & \cellcolor{lightgray!20}{$\mathbf{68.99}$} & \cellcolor{lightgray!20}{$\mathbf{70.33}$} \\
\bottomrule
\end{tabular}}
\end{table*}

\noindent \textbf{Ablation: effect of proposed components.} Table~\ref{ablation:0} shows that all the proposed components, including the frequency encoder, forgery-aware mutual module, and unified decoder, are essential in UFAFormer and bring considerable gains in all sub-tasks. As the frequency domain contains rich face forgery artifacts (Figure~\ref{fig2}) in a manipulated image, it is reasonable that incorporating frequency information can bring a significant improvement for manipulated faces grounding ($+1.49\%$ IoU$\rm_{m}$). Besides, we also observe that frequency information also benefits manipulated words grounding a lot ($+1.58\%$ F1), indicating the enhancement of inconsistent learning between visual and textural modalities.

Without the forgery-aware mutual module (FAMM), we find that the performance of two grounding sub-tasks drops significantly, such as $-3.10\%$ IoU$\rm _{m}$ on face grounding and $-1.27\%$ F1 on word grounding. 
Especially, the faces grounding performance is even worse than without the frequency encoder ($74.72\%$ $vs.$ $76.33\%$ IoU$\rm _{m}$). This verifies the severe problem of {\em semantic conflict} between the image (RGB) and frequency (DWT) features (Section~\ref{sec:FAMM}). The proposed forgery-aware mutual module provides an effective solution to address this problem.

To analyze the unified decoder, we create a variant of our UFAFormer by replacing the decoder with the standard transformer structure \cite{dosovitskiyimage}.
One can see that the unified decoder in UFAFormer is important for all sub-tasks, bringing over $1\%$ improvements on many metrics, {\em e.g.,} $+1.08\%$ AUC on binary classification and $+1.49\%$ mAP on fine-grained manipulation type classification. This is because our unified decoder enables to consolidate visual and textual forgery features and capitalize on the comprehensive relationships among various sub-tasks, thereby easing the optimization and improving the performance.

\noindent \textbf{Ablation: frequency encoder layer implementations.} Table~\ref{ablation:1} provides the comparisons between different frequency encoder layer implementations on two grounding sub-tasks. Specifically, we introduce i) standard self-attentions: we directly concatenate frequency sub-bands and adopt several ViT encoder layers \cite{dosovitskiyimage} to encode mixing features, following \cite{miao2023f}; ii) intra-band self-attentions: we only adopt intra-band interaction among frequency queries; iii) inter-band self-attentions: similar to ii), we solely consider the inter-band interaction; iv) intra-band \& inter-band self-attentions without forgery aggregation: instead of using the prepended sub-band embeddings to perform forgery aggregation to each sub-band, we directly feed the remaining encoded frequency features ($\mathbb{R}^{4\times N\times D}$) to the following module; v) our proposed intra-band \& inter-band self-attentions. 

One can see that our proposed implementation achieves better performance than directly adopting standard self-attention. This is mainly because the forgery artifacts are different in each frequency sub-band (Section~\ref{sec:FE}), while ours explicitly aggregate content and position forgery information within and across different sub-bands.
Besides, we find that both intra-band and inter-band interactions are essential in the frequency encoder layer. Large performance gaps ({\em e.g.,} over $-2\%$ IoU$\rm _{m}$ and $-1\%$ F1) are observed when we remove either of them, as it leads to incomplete interactions among frequency queries in different sub-bands. Moreover, performing forgery aggregation by prepending sub-band embedding is another key to achieving good results. It can be found that the performance of manipulated faces grounding drops $-1.99\%$ without the forgery aggregation. We conjecture that this enables the model to focus on the most discriminative frequency details, avoiding useless noise disturbances such as image backgrounds.

\begin{figure}[t]
\centering
\includegraphics[width=\linewidth]{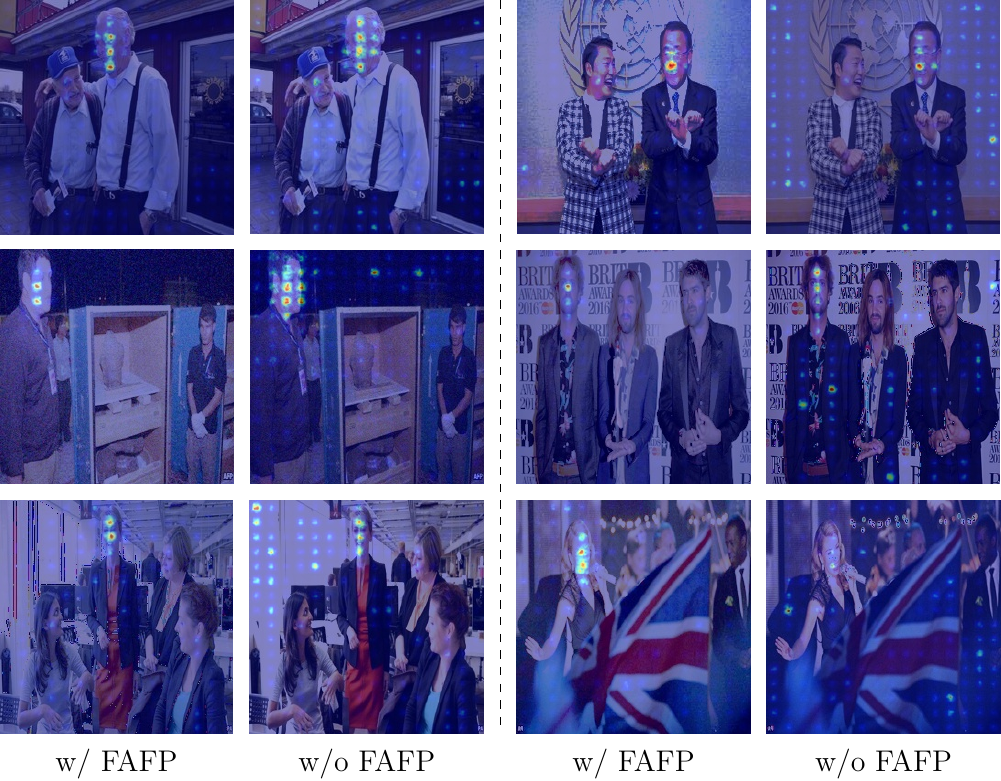}
\caption{\textbf{The gradient norm comparisons of whether adopting forgery-aware feature pyramid (FAFP).} The salient region is visualized by bright color. FAFP looks at forgery face regions, avoiding disturbances from backgrounds. Best view in zoom in.}
\label{fig8}
\end{figure}

\noindent \textbf{Ablation: efficacy of forgery-aware mutual module.}
Table~\ref{ablation:3} evaluates the efficacy of FAMM, including the forgery-aware feature pyramid (FAFP) and mutual cross-attention (MCA). One can see that directly incorporating either FAFP or MCA only provides limited manipulation detecting and grounding improvements. The result gives more verification of the semantic conflict aspects in Section~\ref{sec:FAMM}. 
Besides, we also visualize the gradient norm map with respect to each pixel of given forgery images. The comparisons of with (w/) and without (w/o) FAFP are shown in Figure~\ref{fig8}. The gradient norm reflects the degree of change due to each pixel interference, thus showing which pixels are relied upon for manipulated face detection and grounding. The result shows that FAFP helps UFAFormer to focus on pixels among forgery face regions. With the proposed FAMM, our UFAFormer is able to extract representative and consistent visual forgery features, thus addressing semantic conflicts and obtaining higher performance.

\begin{table}[t]
\renewcommand{\arraystretch}{1.25}
\caption{\textbf{Sampling rate of forgery-aware selection in forgery-aware feature pyramid.} We apply appropriate sampling rates for feature maps with different scales. Larger feature map gets a smaller sampling rate to remove forgery-irrelevant features. `-' denotes not use.}\label{ablation:4}
\resizebox{\linewidth}{!}{
\begin{tabular}{ccccccc}
\toprule
Rate & \multicolumn{3}{c}{Image Grounding} & \multicolumn{3}{c}{Text Grounding} \\\cmidrule{2-7}
(10\%) & IoU$_{\rm m}$ & IoU$_{50}$ & IoU$_{75}$ & PR & RE & F1 \\
\midrule
${\rm -},{\rm -},8$ & $76.92$ & $83.90$ & $77.94$ & $71.44$ & $67.23$ & $69.27$ \\
${\rm -},6,8$ & $77.53$ & $84.91$ & $78.32$ & $71.43$ & $67.86$ & $69.60$ \\
\cellcolor{lightgray!20}{$2,6,8$} & \cellcolor{lightgray!20}{$\mathbf{77.82}$} & \cellcolor{lightgray!20}{$\mathbf{85.15}$} & \cellcolor{lightgray!20}{$\mathbf{78.22}$} & \cellcolor{lightgray!20}{$\mathbf{71.72}$} & \cellcolor{lightgray!20}{$\mathbf{68.99}$} & \cellcolor{lightgray!20}{$\mathbf{70.33}$} \\
$4,8,10$ & $75.40$ & $82.74$ & $74.96$ & $71.45$ & $67.60$ & $69.47$ \\
$10,10,10$ & $74.02$ & $82.40$ & $69.68$ & $70.85$ & $65.16$ & $67.89$ \\
\bottomrule
\end{tabular}}
\end{table}

\noindent \textbf{Ablation: effect of forgery-aware selection.}
Following the forgery-aware selection in the forgery-aware feature pyramid, we tabulate different sampling rate settings for multi-scale feature maps. We apply a smaller sampling rate on a larger feature map to remove forgery-irrelevant features. The results are shown in Table~\ref{ablation:4}. Compared with the setting without selection ({\em i.e.,} `$10,10,10$'), we find that the proposed forgery-aware selection can significantly benefit the performance ($77.82$ $vs.$ $74.02$ AUC and $70.33$ $vs.$ $67.89$ F1). This is mainly because of the identification of forgery-related features, enabling the focus on discerning image manipulation instead of authentic backgrounds. Besides, it can be observed that each scale of feature maps provides gains on image manipulation grounding, such as `${\rm -},6,8$' $vs.$ `${\rm -},{\rm -},8$' brings $+0.61\%$ IoU$\rm _{m}$ and $+0.33\%$ F1. In summary, the sampling rate of $2,6,8$ is enough to extract representative forgery features.

\begin{table}[t]
\renewcommand{\arraystretch}{1.25}
\caption{\textbf{Number of frequency encoder layer.} Increasing layer number can provide gains until the number comes to $3$. Thus, we set $3$ as the default number.}\label{ablation:2}
\resizebox{\linewidth}{!}{
\begin{tabular}{ccccccc}
\toprule
\multirow{2}{*}{\#num} & \multicolumn{3}{@{}c@{}}{Image Grounding} & \multicolumn{3}{@{}c@{}}{Text Grounding} \\\cmidrule{2-7}
& IoU$_{\rm m}$ & IoU$_{50}$ & IoU$_{75}$ & PR & RE & F1 \\
\midrule
$2$ & $77.07$ & $84.87$ & $76.84$ & $71.16$ & $68.20$ & $69.65$ \\
\cellcolor{lightgray!20}{$3$} & \cellcolor{lightgray!20}{$\mathbf{77.82}$} & \cellcolor{lightgray!20}{$\mathbf{85.15}$} & \cellcolor{lightgray!20}{${78.22}$} & \cellcolor{lightgray!20}{$\mathbf{71.72}$} & \cellcolor{lightgray!20}{$\mathbf{68.99}$} & \cellcolor{lightgray!20}{$\mathbf{70.33}$} \\
$4$ & $77.69$ & $84.97$ & $\mathbf{78.30}$ & $71.39$ & $68.66$ & $70.00$ \\
\bottomrule
\end{tabular}}
\end{table}

\noindent \textbf{Ablation: number of frequency encoder layer.} UFAFormer stacks several encoder layers in the frequency encoder. In Table~\ref{ablation:2}, we ablate the effects of layer numbers on binary classification and faces grounding sub-tasks. It can be found that $3$ frequency encoder layers are enough, which achieves similar results with $4$ layers.

\begin{table}[t]
\renewcommand{\arraystretch}{1.2}
\caption{\textbf{Number of grounding embeddings.} When grounding embeddings are greater than $5$, increasing embeddings only leads to limited improvement. Thus, $5$ grounding embeddings are used by default.}\label{ablation:5}
\resizebox{\linewidth}{!}{
\begin{tabular}{ccccccc}
\toprule
\multirow{2}{*}{\#num} & \multicolumn{3}{@{}c@{}}{Binary Class} & \multicolumn{3}{@{}c@{}}{Image Grounding} \\\cmidrule{2-7}
& AUC & EER & ACC & IoU$_{\rm m}$ & IoU$_{50}$ & IoU$_{75}$ \\
\midrule
$3$ & $93.04$ & $14.11$ & $86.46$ & $77.37$ & $84.79$ & $77.89$ \\
\cellcolor{lightgray!20}{$5$} & \cellcolor{lightgray!20}{$\mathbf{93.41}$} & \cellcolor{lightgray!20}{$14.20$} & \cellcolor{lightgray!20}{$\mathbf{86.50}$} & \cellcolor{lightgray!20}{$77.82$} & \cellcolor{lightgray!20}{$\mathbf{85.15}$} & \cellcolor{lightgray!20}{$78.22$} \\
$7$ & $93.39$ & $\mathbf{14.02}$ & $86.25$ & $\mathbf{77.89}$ & $85.10$ & $\mathbf{78.80}$ \\
\bottomrule
\end{tabular}}
\end{table}

\begin{table}[t]
\renewcommand{\arraystretch}{1.2}
\caption{\textbf{Effect of image and text modality.} Both the image (I) and text (T) modality (Mod.) benefit the manipulation detection and grounding.}\label{ablation:6}
\resizebox{\linewidth}{!}{
\begin{tabular}{ccccccc}
\toprule
\multirow{2}{*}{Mod.} & \multicolumn{3}{c}{Image Grounding} & \multicolumn{3}{c}{Text Grounding} \\\cmidrule{2-7}
 & IoU$_{\rm m}$ & IoU$_{50}$ & IoU$_{75}$ & PR & RE & F1 \\
\midrule
I & $74.28$ & $83.47$ & $74.02$ & - & - & - \\
\cellcolor{lightgray!20}{I\&T} & \cellcolor{lightgray!20}{$\mathbf{76.79}$} & \cellcolor{lightgray!20}{$\mathbf{85.15}$} & \cellcolor{lightgray!20}{$\mathbf{77.26}$} & \cellcolor{lightgray!20}{-} & \cellcolor{lightgray!20} {-} & \cellcolor{lightgray!20}{-} \\
\midrule
T  & - & - & - & $73.31$ & $22.96$ & $34.97$ \\
\cellcolor{lightgray!20}{I\&T} & \cellcolor{lightgray!20}{-} & \cellcolor{lightgray!20}{-} & \cellcolor{lightgray!20}{-} & \cellcolor{lightgray!20}{$\mathbf{80.70}$} & \cellcolor{lightgray!20}{$\mathbf{69.00}$} & \cellcolor{lightgray!20}{$\mathbf{74.39}$} \\
\bottomrule
\end{tabular}}
\end{table}

\noindent \textbf{Ablation: number of grounding embeddings.} Table~\ref{ablation:5} gives the ablation of the $K$ randomly initialized learnable grounding embedding number. In UFAFormer, grounding embeddings are utilized to construct image forgery queries, later used for grounding manipulated faces. We observe $K=5$ grounding embeddings are capable to achieve good performance, thus adopting this setting in UFAFormer.

\noindent\textbf{Ablation: effect of each modality.}
In Table~\ref{ablation:6}, we provide more analysis on the effect of each modality in UFAFormer. Specifically, we create two variants by removing either image or text modality from the input and our framework. It can be observed that considering both image and text modalities significantly outperforms other variants, such as $+2.51\%$ IoU$_{\rm m}$ on manipulated faces grounding and $+39.42\%$ F1 on manipulated words grounding. Note that the result of $34.97\%$ F1 in the table is not a bug. The manipulated words by text swap and attribute are hard to identify without reference from the image modality, as the semantic inconsistency in text sequence is very inconspicuous.
The results further demonstrate the effectiveness of multi-modal learning, which enables the framework to leverage cross-modal inconsistency for more accurate manipulation detection and grounding.

\section{Conclusion} \label{sec:conclude}
In this paper, we present a unified transformer-based framework, UFAFormer, for detecting and grounding multi-modal media manipulation. With the well-designed frequency encoder and forgery-aware mutual module, UFAFormer effectively incorporates the DWT frequency domain into the detecting and grounding framework for the first time. Besides, together with the unified decoder, we formulate UFAFormer as a unified and neat framework, setting a new state-of-the-art benchmark on the DGM$^4$ problem. 

\noindent\textbf{Limitations and Future Work.} We notice that our UFAFormer occasionally grounds wrong faces in the presence of severe image perturbation, {\em e.g.,} too smooth Gaussian blur. How to increase the robustness to image perturbation will be further investigated in future works.

\noindent\textbf{Data availability.} The datasets generated during and/or analysed during the current study are available in the DGM$^4$ dataset repository, \url{https://rshaojimmy.github.io/Projects/MultiModal-DeepFake}.


\bibliographystyle{spmpsci}
\bibliography{sample}

\end{document}